\def\year{2021}\relax
\documentclass[letterpaper]{article} 
\usepackage{aaai21}  
\usepackage{times}  
\usepackage{helvet} 
\usepackage{courier}  
\usepackage[hyphens]{url}  
\usepackage{graphicx} 
\usepackage{natbib}  
\usepackage{caption} 
\usepackage[symbol]{footmisc}
\usepackage{tabularx}
\usepackage{threeparttable}
\usepackage{adjustbox}
\usepackage{multirow}
\usepackage{booktabs}
\usepackage[ruled,vlined]{algorithm2e}
\usepackage[dvipsnames]{xcolor}
\usepackage{tablefootnote}
\usepackage[switch]{lineno}

\usepackage{amssymb}

\usepackage{hyperref}
\usepackage{color} 
\usepackage[normalem]{ulem}
\usepackage{multirow}
\usepackage{graphicx}
\usepackage{color}

\def\fig#1{Figure \ref{fig:#1}}

\def\tab#1{Table \ref{tab:#1}}

\def\subsecs#1{\S\ref{subsec:#1}} 

\def\bestviewed{\textit{Best viewed in color.}}

\usepackage{amsmath}
\usepackage{subcaption}

\title{
MAMBA: Multi-level Aggregation via Memory Bank for Video Object Detection
}

\author{
Guanxiong Sun\textsuperscript{\rm 1, \rm 2}, Yang Hua\textsuperscript{\rm 1}, Guosheng Hu\textsuperscript{\rm 2, \rm 1}, Neil Robertson\textsuperscript{\rm 1} \\
}
\affiliations{

    \textsuperscript{\rm 1}EEECS/ECIT, Queen's University Belfast, UK \\
    \textsuperscript{\rm 2}Anyvision, Belfast, UK\\
    \{gsun02, y.hua, n.robertson\}@qub.ac.uk, huguosheng100@gmail.com
}

\begin{document}

\maketitle

\begin{abstract}
State-of-the-art video object detection methods maintain a memory structure, either a \textit{sliding window} or a \textit{memory queue}, to enhance the current frame using attention mechanisms.
However, we argue that these memory structures are not efficient or sufficient because of two implied operations: 
(1) concatenating \emph{all} features in memory for enhancement, leading to a heavy computational cost; (2) \emph{frame-wise} memory updating, preventing the memory from capturing more temporal information. 
In this paper, we propose a multi-level aggregation architecture via memory bank called MAMBA.
Specifically, our memory bank employs two novel operations to eliminate disadvantages of existing methods: (1) \emph{light-weight} key-set construction which can significantly reduce the computational cost;
(2) fine-grained \emph{feature-wise} updating strategy 
which enables our method to utilize knowledge from the whole video. To better enhance features from complementary levels, i.e., feature maps and proposals, we further propose a generalized enhancement operation (GEO) to aggregate multi-level features in a unified manner. 
We conduct extensive evaluations on the challenging ImageNetVID dataset. Compared with existing state-of-the-art methods, our method achieves superior performance in terms of both s
peed and accuracy. More remarkably, MAMBA achieves mAP of 83.7\%$/$84.6\% at 12.6$/$9.1 FPS with ResNet-101. Code is available at \url{https://github.com/guanxiongsun/vfe.pytorch}.

\end{abstract}

\section{Introduction}

\begin{figure}[!t]
  \centering
  \includegraphics[width = \columnwidth]{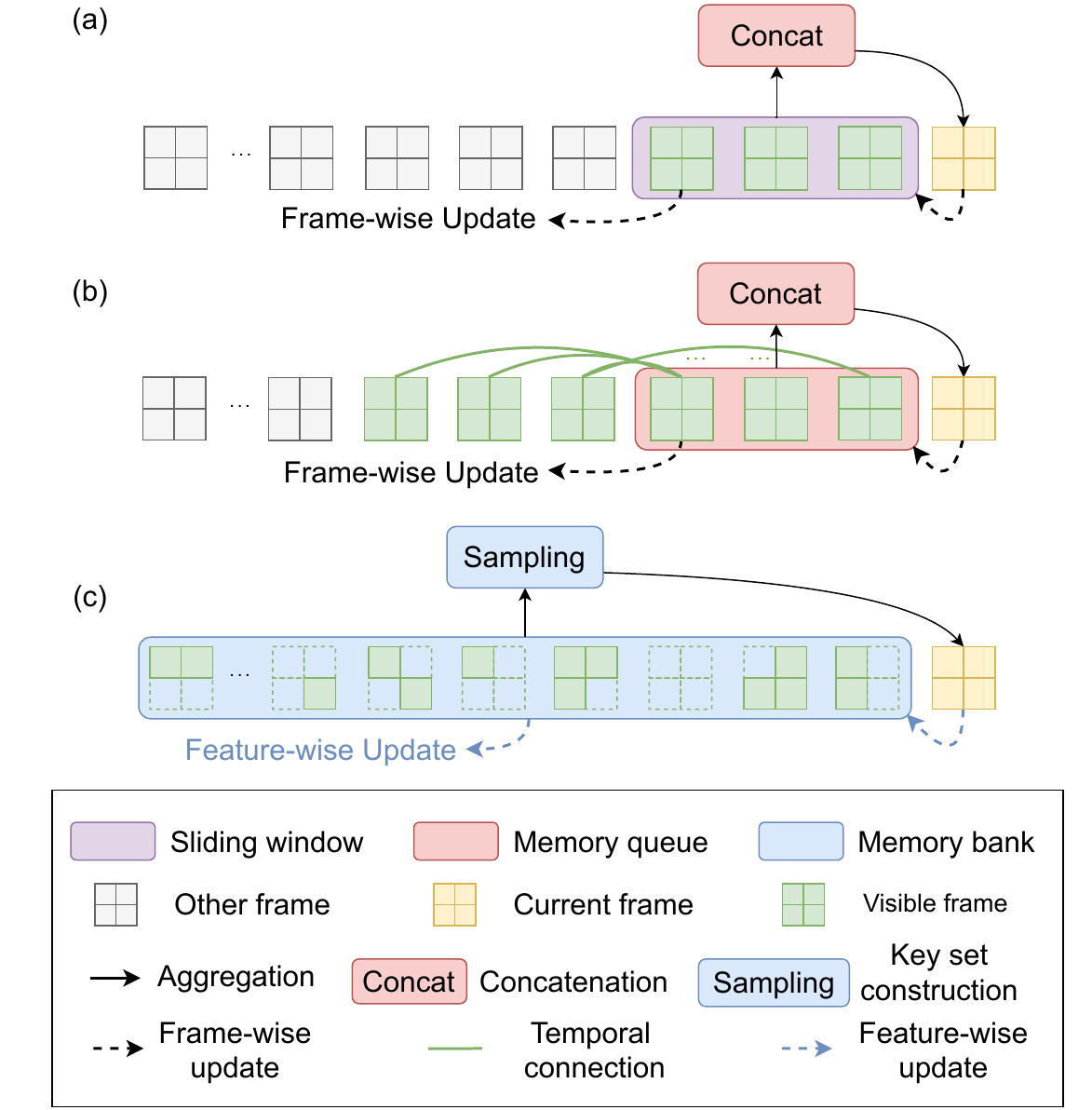}
\caption{Comparisons of the memory construction process in three memory structures.
(a) Sliding window stores raw features of neighbour frames. 
(b) Memory queue stores features of the enhanced frames.
One enhanced frame contains the temporal information of its previous frames. As a result, the number of visible frames is enlarged by temporal connections. 
(c) The proposed memory bank contains two novel operations: light-weight key-set construction and fine-grained feature-wise updating, which help enlarge the number of visible frames to the length of the whole video.
\bestviewed}
\label{fig:motivation}
\end{figure}
Object detection is a fundamental task in computer vision and plays a critical role in many real-world applications. Recently, deep convolutional neural networks (CNNs) based object detectors \cite{rcnn,fastrcnn,fasterrcnn,rfcn,yolo,fcos} have achieved excellent performance on still images. 
However, the success of still-image detectors is hard to transfer to video data directly, because of the quality deterioration of video frames, caused by severe motion blur, rare poses, defocus, occlusions, etc..
To solve these issues, recent methods \cite{dff,thp,fgfa,ogem, selsa, rdn, mega, lrtr} utilize temporal information to enhance video frames, a.k.a., feature-level enhancement methods.
Specifically, feature-level enhancement methods construct a memory structure that contains the features of other frames. Then, either alignment modules, e.g., FlowNet \cite{flownet, flownet2} or relation modules, e.g., attention mechanisms \cite{rn, attention}, are employed to enhance the current frame using features stored in the memory structure.
Depending on how the memory structure is constructed and what features are stored in the memory, existing feature-level enhancement methods can be categorized into two groups: \textit{sliding window} methods and \textit{memory queue} methods.

A \textit{sliding window} \cite{fgfa,manet,stsn,selsa,rdn,lrtr} stores raw features of several neighbour frames of the current frame. Note that the sliding window may contain future frames (offline methods). For demonstration, we show an online version of 
sliding window in \fig{motivation} (a). 
The number of visible frames, which denotes the amount of information the current frame can gather from, is equal to the length of the sliding window.
To enlarge the number of visible frames, \textit{memory queue} methods \cite{mega} utilize recurrent temporal connections to aggregate more temporal information from additional frames, shown in \fig{motivation} (b).
Instead of storing raw features, a memory queue stores the intermediate enhanced features. Thanks to the stacked enhancement stages, a memory queue enlarges the number of visible frames several times, e.g., double in \cite{mega}. 
The increased number of visible frames contributes to better performance.

However, these memory structures are not efficient or sufficient enough because of two implied operations:
(1) Concatenating \emph{all} features in memory for enhancement, leading to heavy computational cost. (2) Updating the memory in a coarse-grained manner, i.e., \emph{frame-wise updating}. Deleting features of the oldest frame at every time step. This limits the number of visible frames to a fixed number, e.g., 20-30, and prevents the model from capturing temporal information from the whole video. 

To address these issues, we propose a \textit{memory bank} for video object detection. As shown in \fig{motivation} (c), our memory bank contains two novel operations.
Firstly, unlike the existing methods \cite{mega,rdn,selsa,ogem} that use \emph{all} the features in the memory, we introduce a light-weight key-set construction strategy to select a subset of features in the memory bank for enhancement, significantly reducing the computational cost and leading to a higher speed.
Secondly, instead of the widely used holistic frame-wise memory updating strategy, we propose a fine-grained feature-wise updating strategy, which can partially delete features from multiple frames. As a result, our method is able to capture and store information from more frames under the same memory size.

In addition, several RFCN-based \cite{rfcn} methods, e.g., MANet \cite{manet} and OGEM \cite{ogem}, demonstrate that the enhancement in different levels, i.e., pixel-level (deep feature maps) and instance-level (position-sensitive score maps), are complementary.
More recent FasterRCNN-based \cite{fasterrcnn} methods \cite{mega,rdn,selsa,lrtr} leverage relation networks \cite{rn} to perform better instance-level enhancements and improve the performance significantly.
However, relation networks cannot receive pixel-level feature as input. To solve this, we introduce a generalized enhancement operation (GEO), which can enhance features in both pixel-level and instance-level in a unified way. By introducing multi-level aggregation via proposed memory bank (MAMBA), our method achieves superior performance in terms of both speed and accuracy. 
To sum up, our contribution is threefold:
\begin{itemize}
	\item We propose a memory bank for video object detection. Specifically, we introduce a light-weight key-set construction strategy and a more fine-grained feature-wise updating mechanism, greatly reducing the computational costs and achieving a flexible framework for different accuracy-speed trade-offs.  
	\item We present a generalized enhancement operation (GEO), which can enhance complementary multi-level (pixel-level and instance-level) features in a unified way. 
	\item We conduct extensive experiments on ImageNet VID dataset \cite{imagenet}. Compared with state-of-the-art methods, our method achieves better performance and faster speed at the same time. 
\end{itemize}

\begin{figure*}[t]
  \centering
  \includegraphics[width = \textwidth]{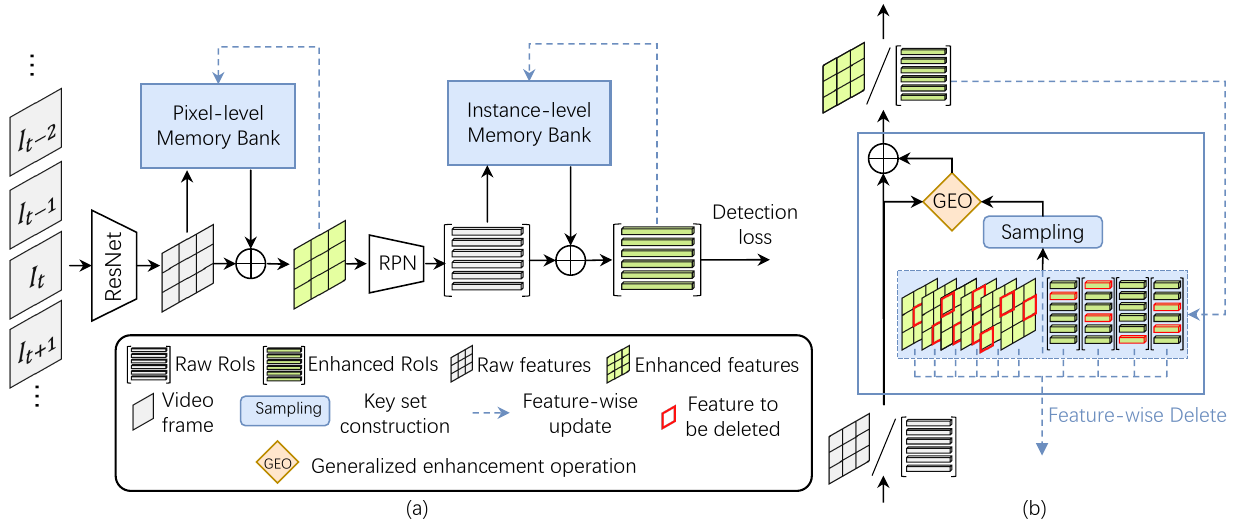}
\caption{
(a) Overview of our framework. Given an input frame $I_t$, firstly, $I_t$ is passed through the backbone networks. Secondly, the extracted feature maps is enhanced by the pixel-level memory bank. Thirdly, the Region Proposal Networks (RPN) is used to extract proposals on the enhanced feature maps. Finally, the proposals are further enhanced by the instance-level memory bank and then enhanced proposals are used to compute the detection loss. (b) Illustration of the enhancement process of the memory bank. The input can either be feature maps or proposals. \bestviewed
}
  \label{fig:method}
\end{figure*}


\section{Related Work}
\noindent\textbf{Object Detection.} Still image object detectors \cite{fastrcnn, fasterrcnn, yolo, yolov2, rfcn} have been remarkably improved due to the development of deep convolutional neural networks (CNNs) \cite{resnet,resnext}. In two-stage detectors \cite{fasterrcnn, rfcn}, firstly, a backbone network \cite{resnet, vgg, inception} is used to extract deep feature maps of an image and then the deep feature maps are passed into the Region Proposal Networks (RPN) \cite{fasterrcnn} to generate object proposals. 
Secondly, the sub-networks further classify the proposals and regress the bounding boxes.
Our proposed memory bank is a general module and can be easily applied to different detectors. 
\\

\noindent\textbf{Video Object Detection.}
Existing video object detection methods can be divided into two categories: box-level methods and feature-level methods. Box-level methods leverage LSTM \cite{tpn} or tracking \cite{tcnn} to model the temporal associations between detected bounding boxes. These methods either introduce heavy computational cost or serve as a post-processing manner \cite{seqnms,d2t}. 
In contrast, the feature-level methods enhance the current frame with other frames end-to-endly.
Based on how to compute the correlation features between frames, feature-level methods can be divided into two subcategories: optical flow based \cite{flownet} and attention \cite{attention} based methods. 
Specifically,
FGFA \cite{fgfa} uses optical flow to align neighbor frames onto the current frame at every time step. THP \cite{thp} does partial aggregation in a recurrent manner. MANet \cite{manet} averages the optical flow within proposals to address the poor flow estimation caused by occlusion.
Recent methods \cite{ogem,selsa,rdn,mega} employ attention mechanisms to do feature enhancement. Most of them \cite{rdn,selsa,mega} conduct instance-level enhancement by reasoning the object relations across frames. OGEM \cite{ogem} proposes an object guided strategy to partially store and sparsely update the object features. 

\section{Our Approach}

\newcommand\mycommfont[1]{\footnotesize\ttfamily\textcolor{Black}{#1}}
\SetCommentSty{mycommfont}

\SetKwInput{KwInput}{Input}                
\SetKwInput{KwOutput}{Output}              

In this section, we introduce a novel framework using Multi-level Aggregation with Memory Bank (MAMBA) for video object detection. In order to enhance the current frame (i.e., query) $Q =\{q_i\}_{i=1}^{N_q}$ (where $N_q$ denotes the number of features in the current frame), we first construct a key set $K=\{k_j\}_{j}^{N_k}$ (where $N_k$ is the total number of feature in the key set) by a light-weight key set construction strategy. The total number of features stored in memory bank $\mathcal{MB}$ is $N_m$ ($N_m$ $\gg$ $N_k$). 
Then, we apply the 
generalized enhancement operation (GEO) to enhance $Q$ with $K$. Note that the proposed GEO supports both pixel-level and instance-level features. Finally, we utilize a feature-wise updating strategy to update memory bank $\mathcal{MB}$. An overview of our framework is shown in Figure \ref{fig:method} and the inference procedure is described in Algorithm 1. 


\subsection{Light-weight Key Set Construction}
\label{subsec:sampling}

Existing methods \cite{mega,ogem,selsa,lrtr,rdn} concatenate \emph{all} the features in memory to build the key set ($K$), which not only increases computational cost but also restricts memory to store more diverse features to further improve the performance. 
In contrast, we design a light-weight key set construction strategy for memory bank to construction the key set, which can be formulated as:
\begin{equation}
    K = Sampling(\mathcal{MB}),
\end{equation}
where $Sampling$ denotes the sampling strategy and $\mathcal{MB}$ denotes memory bank.
Specifically, we implement three sampling strategies: score ranking, frequency select, and random select.
For score ranking strategy, we select the top-$N_k$ features in the memory bank according to their confidence score, either classification score for pixel-level enhancement or objectness score for instance-level enhancement. 
For frequency-guided selection, we normalize confidence scores of all features in the memory bank by softmax. The normalized scores are used as the frequency, which can guide the sampling process. As a result, features with higher scores have higher possibility to be selected, but it is not hard restricted as the score ranking strategy. 
We also implement a random selection strategy. Using random selection, the sampled subset of features approximately follows the same distribution of all features stored in memory. 
In our experiments, all three sampling strategies can effectively 
improve the performance. Among them, random selection achieves a sightly higher accuracy. For simplicity, we use the random selected strategy by default.

It is worth noting that using sampling strategy in the memory bank, we can flexibly control $N_k$ to achieve a controllable speed-accuracy trade-off. In our experiments, we can construct a much larger memory (e.g., $N_m$=96k while the largest $N_m$ for existing methods is only $6k$) to store more diverse features in order to improve performance. Meanwhile, we can select a much smaller key set to achieve a faster speed. 
For example, the smallest $N_k$ used in existing methods is 2,775 reported in RDN \cite{rdn} and RDN achieves 81.8\% mAP with a speed of 128.0 ms. 
An extreme version of our method with $N_k=50$ and $N_m=20,000$ achieves a much higher accuracy of 83.1\% mAP meanwhile a much faster runtime speed at 75.8 ms. Detailed speed-accuracy trade-offs our method are shown in \tab{n_keys}.

\begin{algorithm}[t]
\label{alg:1}
\DontPrintSemicolon
\SetAlgoNoLine
 \caption{Inference Algorithm with Memory Bank in a PyTorch-like style.}
 \BlankLine
 \textcolor{Black}{\#  offline\_test: a bool value denotes whether enable offline testing}
 
 \textcolor{Black}{\#  n\_feat: networks for feature extraction}
 
 \textcolor{Black}{\#  n\_rpn: region proposal networks}
 
 \textcolor{Black}{\#  n\_head: detection head networks for proposals}
 
  \textcolor{Black}{\# GEO: generalized enhancement operation.}
 
 \textcolor{Black}{\# $V$: video frames \{${I_t}$\} of length $T$}
 
 \textcolor{Black}{\# $\mathcal{MB}_{pix}$, $\mathcal{MB}_{ins}$: pixel$/$instance memory bank.}

 \BlankLine
 
 def enhance\_via\_mem\_bank ($q$, $\mathcal{MB}$):
 
 \Indp 
    
    $K$ = $\mathcal{MB}$.sample ()\quad   \textcolor{Black}{\# key set construction}
    
    $\hat{q}$ = $q$ + GEO ($q$, $K$)\quad  \textcolor{Black}{\# generalized enhancement}
 
    return $\hat{q}$
 \BlankLine
 \Indm
 
 if offline\_test: \quad \textcolor{Black}{\# shuffle if do offline testing}
 
 \Indp Random.shuffle($V$) 
 \BlankLine
 
 \Indm
 for $I_t$ in $V$: \quad \textcolor{black}{\# load a frame in the video}
 
 \Indp
 \textcolor{Black}{\# feature extraction networks}
 
 $f$ = n\_feat.forward($I_t$) 
 
 \BlankLine
\textcolor{Black}{\# enhance with pixel-level memory bank}
 
 $\hat{f}_{pix}$ = enhance\_via\_mem\_bank ($f_{pix}$, $\mathcal{MB}_{pix}$)
 
 \BlankLine
 \textcolor{Black}{\# region proposal networks}

 $f_{ins}$ = n\_rpn.forward($\hat{f}_{pix}$) 
 
 \BlankLine
 \textcolor{Black}{\# enhance with instance-level memory bank}
 
 $\hat{f}_{ins}$ = enhance\_via\_mem\_bank ($f_{ins}$, $\mathcal{MB}_{ins}$)
 
 \BlankLine
 \textcolor{Black}{\# detection head networks}
 
 results = n\_head.forward($\hat{f}_{ins}$) 
 
 
 \BlankLine
 \textcolor{Black}{\# feature-wise updating }

 $\mathcal{MB}_{pix}$.update($\hat{f}_{pix}$)
 
 $\mathcal{MB}_{ins}$.update($\hat{f}_{ins}$)
 \BlankLine
 
 
 
\end{algorithm}

\subsection{Unified Multi-level Enhancement }
\label{subsec:enhancement_with_memba}
In this section, we introduce details of the generalized enhancement operation (GEO) which enables multi-level enhancements to be performed in a unified manner. 
Given a query set $Q$ and a key set $K$, the GEO augments each $q_i \in Q$ by measuring $M$ relation features which are achieved by the weighted sum of all key $k$ samples in $K$, where $M$ denotes the number of attention heads. 
Specifically, the $m$-th relation feature $f^{m}_{R}$ of a query sample $q_i$ is calculated as:
\begin{equation}
     f^{m}_{R}(q_i, K) = \sum_j w_{ij}^{m} \cdot (W_{V}^{m} \cdot k_j), \quad m=1, \cdot\cdot\cdot, M,
\end{equation}
where $W^{m}_{V}$ denotes a linear transformation matrix, $M$ is the number of relation features calculated by attention operation and $w_{ij}$ is an element in the correlation matrix $W$ computed based on the similarity of all $q$-$k$ pairs. Precisely, $w_{ij}$ is computed as
\begin{equation}
    w_{ij} = \frac{\exp(S(q_i, k_j))}{\sum_k \exp(S(q_i, k_j)) }, 
\end{equation}
\begin{equation}
    \quad S(q_i,k_j) = \frac{dot(W_Q\cdot q_i, W_K \cdot k_j)}{\sqrt{d}}, 
\end{equation}
where S($q_i$, $k_j$) represents the similarity of $q_i$ and $k_j$, $dot$ denotes the dot product, $W_Q$ and $W_K$ are two transformation matrix, and $d$ is the feature dimension. The total of $M$ relation features are then aggregated by concatenation. Finally, the GEO outputs the augmented feature by adding the original feature $q_i$ and the aggregated relation feature:
\begin{equation}
    GEO(q_i, K) = q_i + concat[{f^{m}_R(q_i, K)}_{m=1}^{M}].
\end{equation}

The enhancement process can be recursively performed. 
Formally, for the $k$-th GEO of enhancement, the augmented feature of $q_i$ is computed as
\begin{equation}
q_i^k = GEO(h(q_i^{k-1}), K),\quad k=1, \cdot\cdot\cdot, N_{g},
\end{equation}
where $h(\cdot)$ denotes the feature transformation function implemented with a fully-connected layer plus ReLU and $N_g$ denotes times of GEO for enhancement recursively. 
With the GEO, we can easily achieve multi-level enhancement, i.e., pixel-level and instance-level feature enhancement, which proves to be effective to utilize complementary feature to further improve the performance.  


\subsection{Feature-wise Updating Strategy}
Existing approaches \cite{rdn,selsa,mega,ogem} update the memory by a frame-wise operation, which deletes all features of the oldest frame.
For the memory bank, we present a fine-grained feature-wise memory updating strategy, which is more flexible and efficient. The feature-wise memory updating strategy can also improve the diversity of features stored in the memory leading to a better performance. 
To implement the feature-wise updating strategy, we use the three sampling methods introduced in \subsecs{sampling} to select features in memory to be updated. 
\subsection{Analysis of Sampling Strategy}
In video object detection, there are many redundant features because adjacent frames are very similar. If the key set has many redundancies, the entropy of information will be small, which means the key set is less informative. The score ranking strategy and frequency selection tend to sample a large portion of features from few frames and decrease the entropy of information. On the contrary, random selection generates a more diverse and informative key set.


\renewcommand{\thefootnote}{\fnsymbol{footnote}}

\section{Experiments}
\begin{table*}[]
\centering
\begin{tabular}{llllc}
\toprule
Methods                                              & Memory & Base detector & Backbone       & mAP(\%) \\
\midrule
D\&T \cite{d2t}                    & -  & RFCN          & ResNet-101     & 75.8    \\
FGFA \cite{fgfa}                    & Window  & RFCN          & ResNet-101     & 76.3    \\
MANet \cite{manet}                  & Window  & RFCN          & ResNet-101     & 78.1    \\
THP \cite{thp}     & Queue  & RFCN          & ResNet-101+DCN & 78.6    \\
STSN \cite{stsn}                    & Window  & RFCN          & ResNet-101+DCN & 78.9    \\
PSLA \cite{psla}   & Queue  & RFCN          & ResNet-101+DCN & 80.0    \\
OGEM \cite{ogem}                                            & Queue  & RFCN          & ResNet-101     & 79.3    \\
\midrule
Ours                                                 & Bank   & RFCN          & ResNet-101     & \textbf{80.8}    \\
\midrule
\midrule
STCA \cite{stca}   & Window  & FasterRCNN    & ResNet-101     & 80.3    \\
SELSA \cite{selsa} & Window  & FasterRCNN    & ResNet-101     & 80.3    \\
LRTR \cite{lrtr}   & Window  & FPN    & ResNet-101     & 81.0    \\
RDN \cite{rdn}     & Window  & FasterRCNN    & ResNet-101     & 81.8    \\
MEGA \cite{mega}   & Queue  & FasterRCNN    & ResNet-101     & 82.9    \\
\midrule
Ours                                                 & Bank   & FasterRCNN    & ResNet-101     & \textbf{84.6}    \\
\midrule
\midrule
LRTR \cite{lrtr}   & Window  & FPN    & ResNeXt-101    & 84.1    \\
RDN \cite{rdn}     & Window  & FasterRCNN    & ResNeXt-101    & 83.2    \\
MEGA \cite{mega}   & Queue  & FasterRCNN    & ResNeXt-101    & 84.1    \\
\midrule
Ours                                                 & Bank   & FasterRCNN    & ResNeXt-101    & 85.4   \\
Ours$^{\dagger}$
& Bank   & FasterRCNN    & ResNeXt-101    & \textbf{86.7}   \\
\bottomrule
\end{tabular}
\caption{Comparison with state-of-the-art end-to-end methods on ImageNet VID validation set. $\dagger$ denotes using random crop and random scale data augmentations for training.}
\label{tab:sota}
\end{table*}

\subsection{Experimental Settings}
\label{sec:exp_setting}
\noindent\textbf{Dataset and Evaluation.} We evaluate our method on 
the ImageNet \cite{imagenet} VID dataset which  contains 3862 training  and 555 validation videos. 
We follow the previous approaches \cite{dff,fgfa,manet,ogem,selsa} and train our model on the overlapped 30 classes of ImageNet VID and DET set. Specifically, we sample 15 frames from each video in VID dataset and at most 2,000 images per class from DET dataset as our training set. Then we report the mean average precision (mAP) on the validation set.



\noindent\textbf{Backbone and Detection Architecture.} Following \cite{dff,fgfa,ogem,rdn} we use the   ResNet-101 \cite{resnet} as our backbone.  Apart from ResNet-101, we also use a stronger backbone ResNeXt-101 \cite{resnext} for some comparisons. For detection network, early methods \cite{dff,fgfa,manet,ogem,stsn} use RFCN \cite{rfcn} as the baseline detector, while more recent methods \cite{selsa,rdn,lrtr,mega} use FasterRCNN \cite{fasterrcnn}. Since our memory bank is a general module and can be applied to different detectors,  we implement memory bank on top of both RFCN and FasterRCNN for fair comparisons. We apply RPN on the extracted deep feature maps. We use 12 anchors with 4 scales ${64^2, 128^2, 256^2, 512^2}$ and 3 aspect ratios ${1:2, 1:1, 2:1}$. Non-maximum suppression (NMS) is applied to generate 300 proposals for each image with an IoU threshold 0.7. Finally, NMS is applied to clean the detection results, with IoU threshold 0.5.

\begin{table*}[t]
\centering
\begin{tabular}{lccccc} 
\toprule
\multirow{2}{*}{Methods} & \multirow{2}{*}{Base detector} & \multirow{2}{*}{mAP(\%)} & \multicolumn{2}{c}{Published}          & Our Impl.       \\
                         &                                &                         & Runtime(ms) & Device                   & Runtime(ms)     \\ 
\midrule
FGFA \cite{fgfa}                     & RFCN                           & 76.3                     & 733         & K40                      & -               \\
MANet \cite{manet}                    & RFCN                           & 78.1                     & 269.7       & Titan X                  & -               \\
OGEM \cite{ogem}                     & RFCN                           & 79.3                     & 112         & 1080 TI                  & 89.1            \\
\midrule
Our$_{pix}$                 & RFCN                           & 80.2                     & \multicolumn{2}{c}{\multirow{2}{*}{-}} & \textbf{81.3}   \\
Our                      & RFCN                           & \textbf{80.8}            & \multicolumn{2}{c}{}                   & 90.1            \\
\midrule
\midrule
STCA \cite{stca}                    & FasterRCNN                     & 80.3                     & 322.2       & Titan X                  & -               \\
SELSA \cite{selsa}         & FasterRCNN                     & 80.3                     & \multicolumn{2}{c}{-}                  & 91.2            \\
RDN \cite{rdn}       & FasterRCNN                     & 81.8                     & 94.2        & V100                     & 128.0           \\
MEGA          \cite{mega}           & FasterRCNN                     & 82.9                     & 114.5       & 2080 TI                  & 182.7           \\
\midrule
Ours$_{ins}$                & FasterRCNN                     & 83.7                     & \multicolumn{2}{c}{\multirow{2}{*}{-}} & \textbf{79.6}   \\
Ours                     & FasterRCNN                     & \textbf{84.6}            & \multicolumn{2}{c}{}                   & 110.3           \\
\bottomrule
\end{tabular}
\caption{Speed-accuracy trade-off with ResNet-101 backbone. The last column shows the runtime(ms) of our implementations. All our results are obtained on Titan RTX GPUs.}
\label{tab:sato}
\end{table*}

\begin{table}[t]
\centering
\begin{tabular}{lcccc}
\toprule
Methods       & Pixel & Instance & mAP(\%)   & \begin{tabular}[c]{@{}c@{}}Runtime\\ (ms)\end{tabular} \\
\midrule
RFCN  &       &          & 73.8      & 46.7                                                   \\
\midrule
Ours$_{pix}$ & \checkmark     &          & 80.2$_{\uparrow 6.4}$  & 81.3                                                   \\
Ours$_{ins}$ &       & \checkmark        & 76.7$_{\uparrow 2.9}$ & 56.0                                                   \\
Ours & \checkmark     & \checkmark        & 80.8$_{\uparrow 6.8}$ & 90.1                                               \\
\midrule
\midrule
FasterRCNN  &       &          & 75.4      & 51.8                                                   \\
\midrule
Ours$_{pix}$ & \checkmark     &          & 81.8$_{\uparrow 6.4}$ & 81.6                                                   \\
Ours$_{ins}$ &       & \checkmark        & 83.7$_{\uparrow 8.3}$ & 79.6                                                   \\
Ours & \checkmark     & \checkmark        & 84.6$_{\uparrow 9.2}$ & 110.3 \\
\bottomrule
\end{tabular}
\caption{ Ablation study of pixel-level and instance-level memory bank on single frame baselines. The first part represents the results using RFCN \cite{rfcn} as the base detector. The second part shows the results using FasterRCNN \cite{fasterrcnn} as the base detector.}
\label{tab:ablation}
\end{table}

\noindent\textbf{Training and Inference Details.} 
To reduce the redundancy and improve the quality of the stored features, we select $K$ samples of enhanced features to update the memory bank. 
Following \cite{rdn,mega}, 
we select $K$=75 proposals with highest objectness score for instance-level memory bank. For pixel-level memory bank, we randomly select $K$=100 pixels within each detected bounding box.
In both training and test phases, the images are resized to a shorter side of 600 pixels. The whole architecture is trained on 4 Titan RTX GPUs with SGD (momentum: 0.9, weight decay: 0.0001).
In the first phase, we only train the pixel-level enhancement. Each GPU contains one mini-batch consisting of two frames, the key frame $I_k$ and a randomly selected frame from the video to approximately form pixel-level memory.  
Both RPN losses and Detection losses 
 are only computed on the key frame. We train the pixel-level model for 60K iterations.
The learning rate is 0.001 for the first 40K iterations, and 0.0001 for the last 20k iterations.
In the second phase, we end-to-end train both pixel-level enhancement and instance-level enhancement for 120K iterations.
The learning rate is 0.001 for the first 80K iterations and 0.0001 for the last 40K iterations.

\subsection{Comparison}

\noindent\textbf{End-to-end performance. } We compare our method with the state-of-the-art methods in
\tab{sota}. 
To make fair comparisons with other methods, we implement our method on top of two base detectors: RFCN \cite{rfcn} and FasterRCNN \cite{fasterrcnn}. 
Table \ref{tab:sota} shows that we achieve the best performance with both RFCN and FasterRCNN setting. Specifically, for RFCN setting, our method also outperforms the best competitor OGEM by 1.5\% mAP and achieve 80.8\% mAP. 
For FasterRCNN setting, our method achieves 84.6\% of mAP and outperforms its best competitor MEGA by 1.7\% using ResNet-101 backbone.
By replacing the backbone with a stronger network ResNeXt-101, our method achieves 85.4\% of mAP. By adding random crop and random scale data augmentations for training, our method finally achieves 86.7\% of mAP.

\noindent\textbf{Speed-accuracy trade-off.} 
To analyze the speed-accuracy trade-off, 
we re-implement many 
state-of-the-art methods and make comparisons 
in Table \ref{tab:sota}.
All results are obtained on Titan RTX GPUs.
Our lite-version model Ours$_{ins}$ which only uses instance-level memory bank achieves both higher accuracy and faster speed than its best competitor MEGA. Specifically, Our$_{ins}$ achieves 83.7\% mAP which outperforms MEGA by 1.7\% mAP. Meanwhile, the speed of Ours$_{ins}$ is 79.6 ms much faster than MEGA. When both pixel-level and instance-level enhancements are performed, the accuracy of our method  is further improved to 84.6\% mAP and the speed is slightly decreased to 110.3 ms.

\subsection{Ablation Study}
To demonstrate the effect of key components in our memory bank, we conduct extensive experiments to study how they contribute to the final performance.

\noindent\textbf{Multi-level enhancement.} In this part, we carefully analyze every component of our method.
\tab{ablation} shows our results using two different base detectors, RFCN (shown in upper rows) and FasterRCNN (shown in lower rows). 
The single frame baseline achieve 73.8\% mAP and 75.4\% mAP for RFCN and FasterRCNN, respectively. 
By introducing the pixel-level memory bank, performances of two baselines are improved to 80.2\% mAP and 81.8\% mAP, respectively.
The improvements introduced by pixel-level enhancement are equal for the two baselines.
By introducing instance-level memory bank, the FasterRCNN baseline is hugely improved by 8.3\% mAP and achieves 83.7\% mAP. However, for the RFCN baseline, the improvement, 2.9\% mAP, is relatively low. 
We believe that the improvement gap is caused by the difference of semantic information. Specifically, the proposals of FasterRCNN have more semantic information than the psroi-pooled features of RFCN.
By utilizing both pixel-level and instance-level memory banks, the performance is further improved to 80.8\% mAP and 84.6\% mAP for the two base detectors. We also show the runtime speed of every model in the last column of Table \ref{tab:ablation}.


\begin{table}[t]
\centering
\begin{tabular}{ccccc}
\toprule
Methods               & (a)  & (b)  & (c)  & (d)  \\
\midrule
Frame-wise updating?   &      & \checkmark    &      &      \\
Feature-wise updating? &      &      & \checkmark    & \checkmark    \\
Class-wise memory?    &      &      &      & \checkmark    \\
\midrule
\midrule
mAP(\%)               & 80.3 & 81.7 & 82.4 & 82.7 \\
\bottomrule
\end{tabular}
\caption{Effect of feature-wise updating strategy and class-wise memory.}
\label{tab:update}
\end{table}

\noindent\textbf{Feature-wise updating.} 
Instead of updating the memory frame-wisely, we propose a feature-wise updating strategy which updates the memory in a more fine-grained manner. 
Specifically, we reuse the three different strategies for key set construction to select features in memory to be deleted. 
The three strategies achieve similar improvements. For simplicity, we use random selection for both key set construction and memory updating in our experiments.
We use SELSA \cite{selsa} as a baseline and denote it as $Model (a)$ in \tab{update}. $Model (b)$ incorporates frame-wise updating (memory queue).
$Model (c)$ incorporates the proposed feature-wise updating strategy.
As shown in Table \ref{tab:update}, comparing with frame-wise updating, feature-wise updating improves the performance by 0.7\% mAP and achieves 82.4\% mAP. 
Previous methods \cite{rdn,mega} only maintain a video-wise memory that deletes all memory by the end of a video. In contrast, we introduce a class-wise memory which is maintained for the whole dataset (multiple videos). In this way, the enhancement process can utilize the information  from other videos. By adding the class-wise memory, $Model (d)$ further improves the performance by 0.3\% to 82.7\% mAP.


\begin{table}[t]
\centering
\resizebox{\columnwidth}{!}{%
\begin{tabular}{lcccc}
\toprule
\multirow{2}{*}{$N_m$} & \multicolumn{2}{c}{Concatenation}        & \multicolumn{2}{c}{Sampling} \\
                      & mAP(\%)           & Runtime(ms)          & mAP(\%)     & Runtime(ms)    \\
                      \midrule
3k                    & 82.3              & 83.4                 & 83.2        & 79.4           \\
6k                    & 82.7              & 92.6                 & 83.4        & 79.3           \\
12k                   & 82.7              & 104.6                & 83.5        & 79.4           \\
24k                   & 82.7              & 142.9                & 83.7        & 79.6           \\
48k                   & \multicolumn{2}{c}{\multirow{2}{*}{OOM}} & 83.7        & 80.3           \\
96k                   & \multicolumn{2}{c}{}                     & 83.8        & 81.0          \\
\bottomrule
\end{tabular}%
}
\caption{Effect of light-weight key set construction with different $N_m$, where $N_m$ denotes the total number of stored samples in the memory bank. OOM denotes out of GPU memory errors on Titan RTX (24GB) devices. }
\label{tab:bank}
\end{table}

\noindent\textbf{Light-weight key set construction.} 
We compare our light-weight key set construction strategy with the widely used concatenation.  
We conduct experiments on $Model (d)$ which concatenates all 6k stored features in memory as the key set. 
We test the overall performance under different number of stored features $N_{m}$,  from $3k$ to $96k$.
For every $N_{m}$, we keep the number of selected keys to $N_{k} = 2k$.
\tab{bank} shows that our light-weight key set construction strategy works always better than concatenation in both speed and accuracy under all $N_{m}$ settings. Using our light-weight key set construction strategy the runtime of the model can always roughly stay the same with the increasing of $N_m$.
In contrast, using concatenation, the speed becomes slower and slower as the increasing of $N_m$. Specifically, when $N_{m}$ is increased to $48k$, concatenation strategy occurs the Out Of Memory (OOM) error on Titan RTX GPU (24GB). We use $N_m=24k$.


\begin{table}[t]
\centering
\begin{tabular}{lccccc} 
\toprule
$N_{k}$  & 50   & 200  & 1k   & 2k*   & 5k    \\
\midrule
mAP(\%)     & 83.1 & 83.5 & 83.6 & 83.7 & \textbf{83.8}  \\
runtime(ms) & \textbf{75.8} & 75.9 & 77.3 & 79.6 & 86.7  \\
\bottomrule
\end{tabular}
\caption{Analysis of different number of $N_{k}$. }
\label{tab:n_keys}
\end{table}

\begin{table}[t]
\centering
\begin{tabular}{lcccc}
\toprule
$N_{pix}$        & 0    & 1*    & 2   &  3 \\
\midrule
mAP(\%)      & 75.4 & 81.8 & 81.8  & 81.7 \\
Runtime (ms) & 51.8 & 81.6 & 112.2 & 143.1 \\
\bottomrule
\end{tabular}
\caption{Analysis of different number of $N_{pix}$.}
\label{tab:n_pixel}
\end{table}

\begin{table}[t]
\centering
\begin{tabular}{lllll}
\toprule
$N_{ins}$        & 0    & 1          & 2*    & 3          \\
\midrule
mAP(\%)      & 75.4 & 82.0 & 83.7 & 83.8 \\
Runtime (ms)   & 51.8 & 65.1 & 79.6 & 94.1  \\
\bottomrule
\end{tabular}
\caption{Analysis of different number of $N_{ins}$. }
\label{tab:n_instance}
\end{table}

\begin{table}[!t]
\centering
\begin{tabular}{clll}
\toprule
Pixel & AR$^5$  & AR$^{10}$ & AR$^{100}$  \\
\midrule
            & 77.9 & 83.8 & 94.3  \\
\checkmark   & \textbf{79.5$_{\uparrow 1.6}$}  & \textbf{85.7$_{\uparrow 1.9}$}  & \textbf{96.3$_{\uparrow 2.0}$}   \\
\bottomrule
\end{tabular}
\caption{Effect of pixel-level enhancement to RPN.}
\label{tab:rpn}
\end{table}

\noindent\textbf{Size of the key set.} We evaluate how the size of the key set $N_{k}$ effects the performance.
In this part, we conduct experiments on the our lite-version method Ours$_{ins}$ which only performs two times of instance-level enhancement $N_{ins}=2$ and no pixel-level enhancement $N_{pix}=0$. 
From Table \ref{tab:n_keys}, 
our method is very robust to the number of sampled keys. In practice, we set $N_{k}=2000$ for better speed-accuracy trade-off.
Surprisingly, even sampling a very small number of keys $N_k=50$, 
our method still achieves 83.1\% mAP and outperforms the best competitor MEGA by 0.2\% mAP. 


\noindent\textbf{Number of pixel-level enhancements.}
As discussed in \subsecs{enhancement_with_memba}, the enhancement with memory bank can be performed multiple times. We evaluate the effect of the number of pixel-level enhancements $N_{pix}$.
When $N_{pix}=0$, no pixel-level enhancement is performed, our method degenerates to the FasterRCNN baseline.
In Table \ref{tab:n_pixel}, we vary $N_{pix}$ from 0 to 3. With $N_{pix}=1$,
the performance is improved by 6.4\% to 81.8\% mAP. By further increasing $N_{pix}$, the performance is merely improved. We use $N_{pix}=1$ by default.

\noindent\textbf{Number of instance-level enhancements. } Similarly, we evaluate the effect of the number of instance-level enhancements $N_{pix}$ by performing instance-level enhancement on top of the FasterRCNN. When $N_{ins}=0$, no instance-level enhancement is performed. 
From Table \ref{tab:n_instance}, we can see that $N_{ins}=2$ achieves the best speed-accuracy trade-off. Specifically, the performance is improved by 8.3\% to 83.7\% mAP with $N_{ins}=2$. We use $N_{ins}=2$ by default.

\noindent\textbf{Effect of pixel-level memory bank to RPN.} Recent instance-level methods \cite{mega,selsa,rdn} do not enhance deep feature maps used in RPN. 
Thus, the RPN potentially misses some low-quality objects.
We  evaluate the effect of pixel-level memory bank to RPN. Specifically, the metric Average Recall (AR) is used for comparison. 
We select top $k$ of the proposals generated by RPN to calculate the AR$^k$. Specifically, we tested with $k=\{5,10,100\}$. As shown in Table \ref{tab:rpn}, with pixel-level enhancement, AR$^5$, AR$^{10}$, and AR$^{100}$ are all improved. 


\section{Conclusions}
In this paper, we propose a multi-level aggregation framework via memory bank (MAMBA). The memory bank contains two novel operations: (1) light-weight key-set construction and (2) fine-grained feature-wise memory updating. 
Experiment results demonstrate MAMBA achieves superior performance on the challenging ImageNet VID dataset in terms of both speed and accuracy. 

\newpage
\bibliography{egbib}

\end{document}